\documentclass[letterpaper]{article} 
\usepackage{aaai2027}  
\usepackage[hyphens]{url}  
\usepackage{graphicx} 
\usepackage{natbib}  
\usepackage{caption} 
\usepackage{algorithm}
\usepackage{algorithmic}
\usepackage{amsmath}

\usepackage{newfloat}
\usepackage{listings}
\DeclareCaptionStyle{ruled}{labelfont=normalfont,labelsep=colon,strut=off} 
\floatstyle{ruled}
\newfloat{listing}{tb}{lst}{}
\floatname{listing}{Listing}

\usepackage{booktabs}

\nocopyright

\title{Beyond Pixel Reconstruction: Retrieval-Guided Glyph-Aware Restoration for Low-Resource Manchu Historical Documents}
\author{
    Ting Huang\textsuperscript{\rm 1,*},
    Dongdong Wang\textsuperscript{\rm 2,*},
    Mingqiu Liang\textsuperscript{\rm 1},
    Siyang Lu\textsuperscript{\rm 1,$\dagger$}
}

\affiliations{
    \textsuperscript{\rm 1}School of Computer Science and Technology, Beijing Jiaotong University, Beijing, China\\
    \textsuperscript{\rm 2}College of Design, Construction, and Planning, University of Florida, Gainesville, FL, USA\\
}

\begin{document}

\maketitle
\begingroup
\renewcommand{\thefootnote}{\fnsymbol{footnote}}
\footnotetext[1]{Equal contribution.}
\footnotetext[2]{Corresponding author: Siyang Lu (sylu@bjtu.edu.cn).}
\endgroup

\begin {abstract}
Restoring low-resource historical documents, such as Manchu historical documents, is critical for preserving cultural heritage but remains challenging due to severe data scarcity and complex degradation patterns. Although recent diffusion-based restoration methods have substantially improved visual reconstruction quality, the scarcity of annotated Manchu historical documents necessitates transferring pretrained models from natural-image or other writing-system domains. Such transferred models provide limited script-specific knowledge, while their pixel-centric optimization objectives fail to explicitly preserve legitimate Manchu glyphs, let alone recover the correct ones. We tackle this challenge by reformulating historical document restoration from pixel reconstruction to glyph-aware recovery, leveraging a script-specific glyph inventory together with the semantic reasoning capability of a large vision-language model, rather than directly synthesizing missing pixels. Accordingly, we propose a Coarse-to-Fine Retrieval-Guided Glyph-Aware Restoration (CRGR) framework, in which a vision encoder retrieves structurally similar candidates and a large vision-language model identifies the most appropriate glyph. This retrieval-before-restoration paradigm enables glyph-aware restoration without task-specific training while promoting the recovery of legitimate Manchu glyphs. We further establish a \textbf{\emph{Glyph-oriented Evaluation Protocol}}, demonstrating that faithful restoration should be evaluated beyond image fidelity by explicitly assessing glyph validity, identity preservation, and structural consistency. Extensive experiments demonstrate that CRGR achieves the highest glyph structural similarity (GSS = 0.9986) and superior restoration accuracy on intact content (GRA = 0.9600), validating retrieval-before-restoration as an effective paradigm for improving glyph fidelity in low-resource Manchu historical document restoration.
\end {abstract}


\section{Introduction}

Historical document restoration is a fundamental task in cultural heritage preservation and digital humanities, aiming to recover degraded historical documents for subsequent archival conservation, scholarly analysis, and digital dissemination \citep{Pal2017}. Recent advances in deep learning have significantly improved historical document restoration across various degradation scenarios \citep{Zhu2024}. However, the effectiveness of existing methods largely depends on the characteristics of target writing systems, including training data availability, glyph diversity, and document degradation patterns. Existing restoration methods have achieved promising results on high-resource scripts,where relatively large-scale datasets and representative training samples are available. In contrast, many under-resourced writing systems, including Manchu, Khitan, and Jurchen, contain only limited historical collections and exhibit substantial variations in glyph styles. Furthermore, long-term aging, insect damage, ink fading, and physical deterioration introduce complex degradation patterns, making restoration particularly challenging for low-resource scripts. Consequently, restoration failures frequently arise not only from missing visual details but also from inaccurate recovery of glyph structures.

\begin{figure*}[t]
\centering
\includegraphics[width=\linewidth]{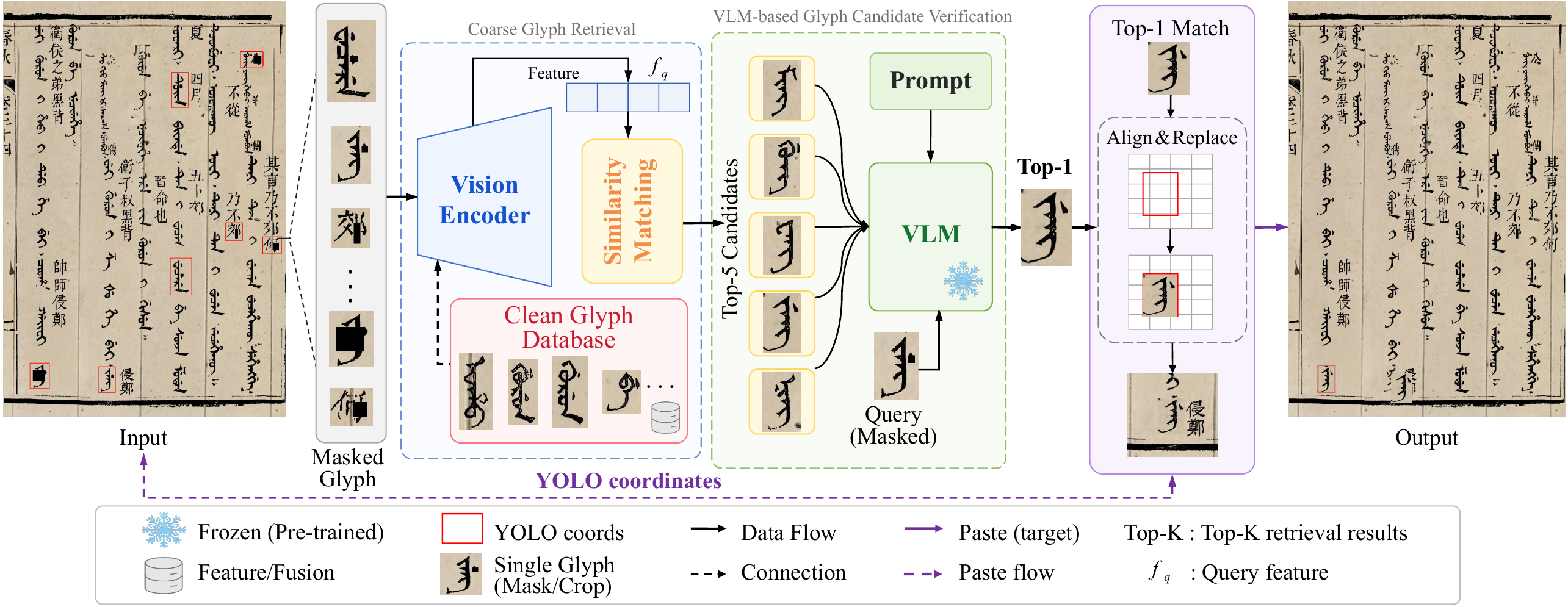}
\caption{Overview of the proposed CRGR framework. CRGR reformulates restoration from pixel synthesis to retrieval-guided glyph recovery by progressively reducing candidate ambiguity through coarse retrieval and VLM-based verification.}
\label{fig:cgvm_framework}
\end{figure*}

Existing restoration methods mainly formulate historical document restoration primarily as a pixel reconstruction problem. Their objective is to infer missing contents from surrounding image contexts and generate visually plausible results \citep{Suvorov2022}. This pixel-centric formulation implicitly assume that visual similarity is a reliable surrogate for glyph correctness. However, this assumption becomes unreliable for low-resource historical scripts containing highly similar glyph structures. Achieving robust restoration under this paradigm typically requires abundant annotated data, powerful pretrained models, or task-specific adaptation, which are difficult to obtain for many endangered writing systems. Consequently, visually plausible reconstructions may still contain incorrect glyph structures or distorted stroke patterns. Since restoration errors can propagate to downstream tasks such as optical character recognition, historical interpretation, and linguistic analysis, preserving glyph structures and visual integrity is more important than pixel-level similarity alone. 

To address these challenges, we reformulate historical document restoration as a retrieval-guided glyph recovery problem and propose Coarse-to-Fine Retrieval-Guided Glyph-Aware Restoration (CRGR), taking bilingual Manchu--Chinese woodblock-printed historical documents as a representative low-resource restoration benchmark. CRGR first employs a vision encoder to retrieve Top-$K$ structurally similar glyph candidates from an intact glyph gallery, followed by fine-grained candidate verification using a frozen vision-language model. Directly applying VLM verification over the entire glyph inventory can introduce substantial computational overhead, CRGR progressively reduces the candidate space before multimodal verification. The selected reference glyph is then restored through reference-guided replacement. By combining coarse retrieval with semantic verification, CRGR mitigates ambiguities among structurally similar glyphs without requiring task-specific fine-tuning. This formulation shifts historical document restoration from open-ended pixel synthesis toward constrained reference selection, reducing the risk of hallucinating invalid glyph forms.

The main contributions are summarized as follows:
\begin{itemize}
    \item We reformulate historical document restoration from pixel synthesis to structure-aware glyph reference selection. Based on this formulation, we propose CRGR, which integrates visual retrieval and frozen VLM verification to achieve glyph-preserving restoration without task-specific fine-tuning.

    \item We develop a hierarchical candidate refinement strategy that bridges coarse glyph retrieval and fine-grained VLM verification. By progressively reducing candidate ambiguity, CRGR enables reliable identification among structurally similar glyphs.

    \item We establish a glyph-oriented evaluation protocol that extends restoration assessment beyond pixel-level fidelity by explicitly measuring glyph identity preservation and structural consistency through GRA, CCR, and GSS.

    \item Extensive experiments on a Manchu historical document dataset demonstrate that CRGR achieves superior glyph fidelity and document content preservation compared with representative restoration methods, validating the effectiveness of the proposed retrieval-guided restoration paradigm.
\end{itemize}

\begin{table*}[t]
\centering
\small
\setlength{\tabcolsep}{3.2pt}
\begin{tabular}{l l l l}
\toprule
\textbf{Stage} & \textbf{Input} & \textbf{Transformation} & \textbf{Output} \\
\midrule
\textbf{Coarse Glyph Retrieval}
& Masked glyph $x_q$, gallery $\mathcal{G}$
& Visual feature extraction + retrieval
& Top-$K$ glyph candidates $\mathcal{C}$ \\

\textbf{Candidate Verification}
& Masked glyph $x_q$, candidates $\mathcal{C}$
& VLM-based vision-language matching
& Reference glyph $g^{*}$ \\

\textbf{Glyph Restoration}
& Reference glyph $g^{*}$, target location
& Glyph alignment and regional replacement
& Restored documents \\
\bottomrule
\end{tabular}
\caption{Data transformation across different stages of the proposed CRGR framework.}
\label{tab:pipeline_data}
\end{table*}

\section{Related Work}

\subsection{Visual Reconstruction}
Most existing historical glyph restoration methods formulate restoration as a visual reconstruction problem, where missing contents are inferred from surrounding image contexts and optimized for pixel-level fidelity \cite{Pathak2016,Yu2019}. Early deep learning approaches mainly rely on context-aware image inpainting to recover visually plausible contents, represented by adaptive attention-based models such as AOT-GAN \citep{Zeng2021}. To better handle large missing regions, large-receptive-field restoration methods, such as LaMa \citep{Suvorov2022}, exploit Fourier convolutions to capture long-range contextual dependencies and improve structural continuity. More recently, diffusion-based restoration methods, such as Unified Diffusion Transformer (UDT) \citep{Kim2025}, have demonstrated remarkable generation capability by progressively refining corrupted regions through iterative denoising. Meanwhile, document-oriented restoration frameworks, such as TextDoctor \citep{Lu2025}, DocRevive \citep{Purkayastha2026}, and GSDM \citep{Zhu2024}, further incorporate textual structures, degradation-aware priors, and document layouts to improve restoration quality on degraded historical documents. Despite their strong visual reconstruction capability, these methods primarily optimize appearance consistency and implicitly assume that visually plausible restoration preserves glyph structures. This assumption becomes unreliable for low-resource historical scripts containing visually similar glyph structures, where visual realism does not necessarily guarantee structural correctness.

\subsection{Structure-aware Restoration}
To improve glyph integrity during restoration, recent studies introduce explicit structural constraints beyond conventional image reconstruction. Stroke-aware restoration methods preserve character topology by modeling stroke continuity and structural dependencies, while Transformer-based approaches, such as LWTFormer \citep{Ruan2026}, further capture long-range relationships among character components. Diffusion-based structural restoration has also been explored for ancient Han characters through explicit glyph generation constraints \citep{Li2024}. In parallel, adversarial learning remains an effective solution for structure-preserving restoration. Representative examples include Three-Stage GAN \citep{Shen2026}, which progressively reconstructs broken strokes, and Dual Discriminator GAN \citep{Chen2022}, which jointly constrains global appearance and local structural consistency. Another important direction is exemplar-guided restoration, where visually similar reference glyphs are introduced to guide reconstruction, as demonstrated by EA-GAN \citep{Zheng2022}. Although these methods significantly improve structural fidelity and glyph completeness, restoration decisions are still dominated by visual similarity. Structurally similar glyphs remain difficult to distinguish based solely on visual appearance, particularly for low-resource historical scripts with limited training data. Preserving glyph structures alone cannot address the challenge of distinguishing visually similar glyph patterns.

\subsection{Retrieval-guided and Vision-language Restoration}
Recent studies have begun moving beyond purely pixel-level image generation by incorporating glyph retrieval and vision-language models to improve glyph discrimination during historical document restoration. These methods can be broadly categorized according to how glyph retrieval and vision-language reasoning are incorporated into the restoration process. Early efforts focused on learning more discriminative glyph representations. Restoring Ancient Ideograph jointly optimizes image restoration, character recognition, and character prediction within a unified multi-task framework, enabling more discriminative glyph feature learning \citep{Duan2024}. Subsequent studies introduced cross-modal representation alignment to jointly model visual glyph structures and auxiliary textual information for historical inscription restoration \citep{Liu2025}. Benefiting from recent advances in multimodal learning, large vision-language models have also been introduced into historical document restoration. EpiAgent \citep{Zhu2026} introduces multimodal reasoning for historical document understanding, while UniHIR \citep{Zhang2026} further integrates unified multimodal modeling for historical inscription restoration. Despite these advances, most existing methods rely on abundant annotated data, task-specific adaptation, or end-to-end multimodal optimization. Many VLM-augmented restoration approaches aim to synthesize pixel content directly and require domain fine-tuning. Such requirements are difficult to satisfy for low-resource historical documents. However, existing methods generally do not explicitly decouple candidate retrieval from candidate verification. Consequently, existing approaches either optimize pixel generation or rely on task-specific adaptation, while retrieval-guided semantic restoration utilizing authentic glyph references has not been sufficiently explored for low-resource historical documents.

\section{Methodology}
\subsection{Framework Overview} 
Following the retrieval-guided formulation, CRGR consists of two consecutive stages: coarse glyph retrieval and candidate verification. The first stage retrieves a compact set of structurally similar glyph candidates using a vision encoder, while the second stage performs fine-grained candidate verification using a frozen vision-language model (VLM) to identify the most appropriate reference glyph \cite{Radford2021}. The selected reference glyph is then inserted into the corrupted region to complete document restoration. The overall architecture is illustrated in Fig.~\ref{fig:cgvm_framework}, and the data flow of the proposed framework is summarized in Table~\ref{tab:pipeline_data}.

\subsection{Glyph Knowledge Base}

Unlike generative restoration methods that synthesize missing content from learned priors, our framework restores damaged glyphs by retrieving valid references from an authentic glyph knowledge base. The knowledge base is constructed offline from intact historical historical documents and serves as the search space for subsequent retrieval and verification.

Specifically, each intact glyph is first localized and cropped using a pretrained glyph detector. The extracted glyphs are then normalized to a unified resolution while preserving their glyph identities. Collectively, these glyph instances form a reference gallery
$\mathcal{G}=\{g_1,g_2,\ldots,g_N\}$, where each $g_i$ denotes an authentic glyph observed in historical documents.

The knowledge base is fixed during inference and is shared across all restoration samples. Since every restored glyph is selected from $\mathcal{G}$ rather than synthesized pixel by pixel, the proposed framework naturally constrains restoration to valid historical glyphs, reducing the possibility of generating unsupported glyph forms.

\subsection{Coarse Glyph Retrieval}
Given a damaged glyph, the first stage of CRGR retrieves a compact set of structurally similar reference glyphs from an intact glyph gallery using discriminative visual representations. A vision encoder is employed to extract hierarchical glyph features from both damaged glyphs and intact reference glyphs. These representations encode hierarchical visual patterns of glyph structures, providing discriminative features for retrieval.

For a damaged glyph $x_q$, its feature representation is denoted as $f_q = E(x_q)$, where $E(\cdot)$ represents the vision encoder. Likewise, each reference glyph $g_i$ in the glyph gallery is encoded as $f_i = E(g_i)$.

The similarity between the damaged glyph and each reference glyph is measured using cosine similarity,
\begin{equation}
s_i = \frac{f_q^\top f_i}{\|f_q\| \|f_i\|}.
\label{eq:sim}
\end{equation}

All reference glyphs are ranked according to their similarity scores, and the Top-$K$ candidates are retained to form the candidate set,
\begin{equation}
\mathcal{C}
= \text{TopK}\big(\text{rank}(s_i), K\big).
\label{eq:candidate_set}
\end{equation}

The Top-$K$ candidates are preserved for subsequent VLM-based verification instead of selecting the highest-ranked result. This design substantially reduces the search space while ensuring that every restored glyph originates from an authentic historical reference. Moreover, restricting restoration to retrieved historical references naturally prevents generating glyph forms that do not exist in the writing system.

\subsection{VLM-based Candidate Verification}
Visual similarity alone is insufficient because structurally similar glyphs may still exhibit subtle structural differences that are difficult to distinguish. The VLM compares the damaged glyph with each retrieved candidate under a predefined prompt and estimates their matching scores. The candidate with the highest matching score is selected as the reference glyph for restoration. The VLM receives the masked glyph and each retrieved candidate as input and outputs a matching score under a fixed prompt template.


Given the candidate set $\mathcal{C}$ retrieved from the reference gallery $\mathcal{G}$, the VLM assigns a verification score to each candidate glyph $g_i\in\mathcal{C}$:
\begin{equation}
s_i^{\mathrm{vlm}} = V(x_q, g_i),
\end{equation}
where $V(\cdot)$ denotes the VLM-based verification function and $x_q$ is the query. The final reference glyph is selected as
\begin{equation}
g^*=\arg\max_{g_i\in\mathcal{C}} s_i^{\mathrm{vlm}}.
\label{eq:final_glyph}
\end{equation}

The selected reference glyph is subsequently aligned to the target position and reference-guided replacement to complete restoration. The two-stage design combines efficient candidate retrieval with fine-grained verification, enabling reliable glyph selection without task-specific fine-tuning.

\begin{figure*}[t]
\centering
\includegraphics[width=\linewidth]{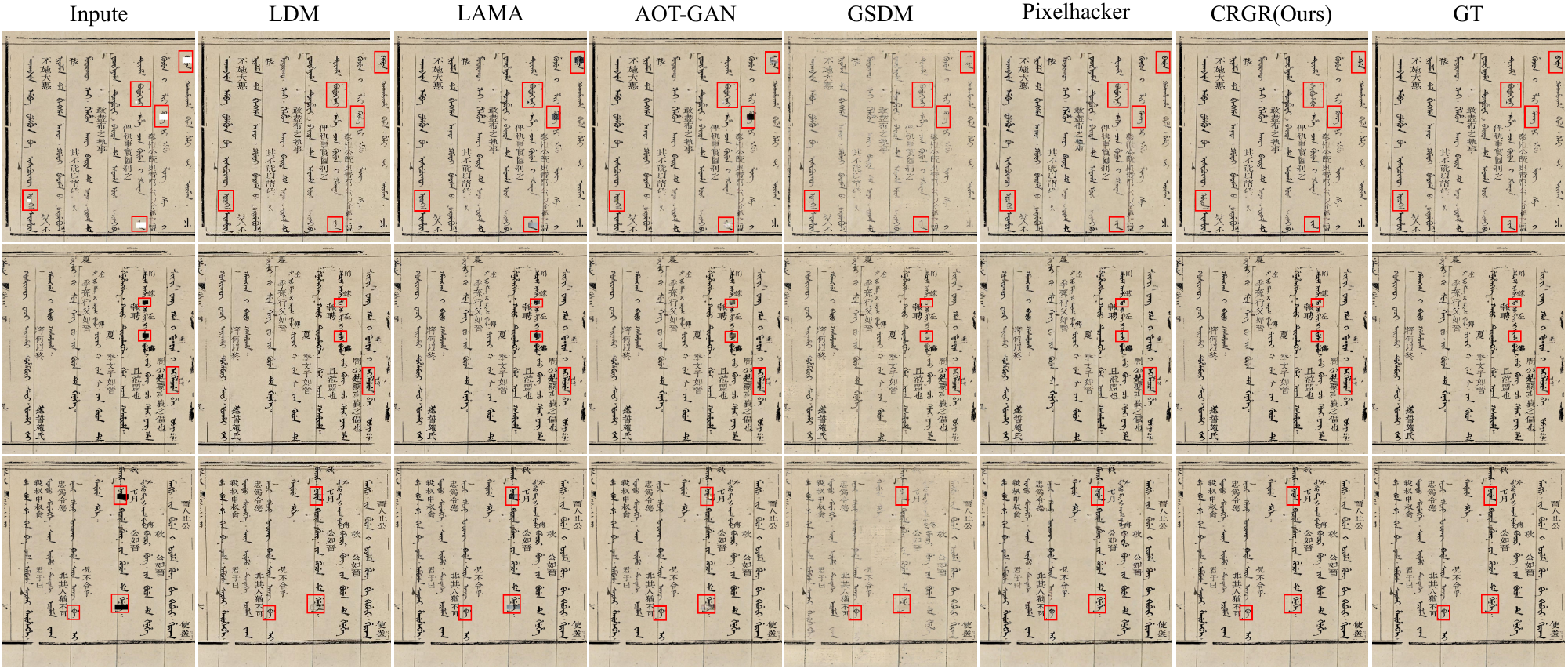}
\caption{Qualitative comparison of full-page document restoration.}
\label{fig:full_page_compare}
\end{figure*}

\subsection{Glyph-oriented Evaluation Protocol}

Existing restoration metrics, such as PSNR, SSIM, and LPIPS, primarily assess image fidelity and cannot distinguish structurally similar glyphs with different semantic identities. As a result, visually plausible but semantically incorrect restorations may still receive high level scores. To address this limitation, we establish a \textbf{\emph{Glyph-oriented Evaluation Protocol}} that explicitly evaluates restoration quality from the perspective of glyph correctness rather than pixel similarity.

Our protocol consists of two stages. First, a pretrained glyph detector localizes individual glyph instances from restored historical documents and extracts discriminative glyph representations. Second, the detected glyphs are evaluated using three complementary metrics that jointly assess glyph identity preservation and structural fidelity: Glyph Restoration Accuracy (GRA), Cross-category Confusion Rate (CCR), and Glyph Structure Similarity (GSS).

GRA measures the proportion of correctly restored glyph identities:
\begin{equation}
\mathrm{GRA}=\frac{N_{\mathrm{correct}}}{N_{\mathrm{total}}},
\label{eq:gra}
\end{equation}
where $N_{\mathrm{correct}}$ and $N_{\mathrm{total}}$ denote the numbers of correctly restored glyphs and evaluated glyphs, respectively.

CCR quantifies semantic confusion between visually similar glyph categories:
\begin{equation}
\mathrm{CCR}=\frac{N_{\mathrm{cross}}}{N_{\mathrm{total}}},
\label{eq:ccr}
\end{equation}
where $N_{\mathrm{cross}}$ denotes the number of glyphs restored as an incorrect category.


To further evaluate structural fidelity after identity verification, we define GSS based on deep glyph features. Given the feature extractor $\phi(\cdot)$ and the intact glyph gallery $\mathcal{G}$, GSS is computed as
\begin{equation}
\mathrm{GSS}
=
\frac{1}{N_{\mathrm{correct}}}
\sum_{i=1}^{N_{\mathrm{correct}}}
\max_{g_j\in\mathcal{G}}
\frac{\phi(r_i)^\top\phi(g_j)}
{\|\phi(r_i)\|_2\,\|\phi(g_j)\|_2},
\label{eq:gss}
\end{equation}
where $\phi(\cdot)$ denotes the feature extractor used for structural comparison,
and $r_i$ denotes the $i$-th restored glyph.

Unlike conventional image-level evaluation, the proposed protocol explicitly separates glyph identity correctness from structural similarity, enabling faithful assessment of restoration quality at the semantic glyph level. Since glyph localization and representation learning are modular components, the protocol is detector-agnostic and can readily incorporate future advances in glyph detection and feature extraction.

\section{Experiments}
\subsection{Experimental Setup}
\subsubsection{Dataset.}
To the best of our knowledge, no public benchmark targets retrieval-guided glyph restoration; therefore, we construct a benchmark of scanned Manchu-Han bilingual woodblock-printed documents from multiple sources. We evaluate CRGR under full-page and patch-level restoration settings. For patch-level evaluation, each page is divided into $3\times3$ sub-regions, and patches containing masked glyphs are retained.
A pretrained YOLO detector constructs an intact glyph gallery, yielding over 15,000 reference glyph instances. The gallery serves as the retrieval database for CRGR.

\subsubsection{Baselines.}
We compare CRGR with five representative restoration approaches covering different restoration paradigms, including CNN-based, GAN-based, diffusion-based, and document restoration methods. We select publicly available methods with released pretrained models and reproducible inference pipelines. Following the official implementation setting, LDM adopts a $512\times512$ input resolution for full-page restoration \cite{Rombach2021}.

\subsubsection{Evaluation Metrics.}
We primarily evaluate restoration performance using the proposed glyph-oriented metrics, including Glyph Restoration Accuracy (GRA), Cross-category Confusion Rate (CCR), and Glyph Structure Similarity (GSS). These metrics are reported on both masked and intact regions to evaluate glyph recovery and preservation of original document contents. In addition, image-level metrics, including PSNR, SSIM, LPIPS, RMSE, and F1-score for foreground reconstruction, are additionally reported.

\subsubsection{Implementation Details.}
The proposed CRGR consists of a coarse glyph retrieval stage and a VLM-based candidate verification stage. We adopt Swin V2 as the visual encoder for glyph feature extraction and candidate retrieval, and employ a frozen InternVL model for candidate verification \cite{Liu2021, Chen2024}. The VLM receives the damaged glyph and each retrieved candidate as input and produces a matching score under a fixed prompt template. The candidate with the highest verification score is selected from the Top-$K$ retrieved candidates as the reference glyph. We set $K=5$ in all experiments.
All experiments are conducted on a single NVIDIA RTX 4090 GPU (24 GB) using the PyTorch framework. All backbone models utilize pretrained weights without task-specific fine-tuning. Ablation studies further analyze the effects of different retrieval backbones, VLM verification models, and candidate numbers.

\subsection{Comparison with Representative Restoration Methods}
\subsubsection{Qualitative Comparison.}
We first compare CRGR with representative restoration methods under the same masking setting. Figure~\ref{fig:full_page_compare} presents qualitative comparisons on full-page document restoration. Specifically, LaMa and AOT-GAN may produce noticeable artifacts in masked regions, including local appearance inconsistencies and distorted stroke patterns, possibly due to the absence of explicit glyph-level constraints. Diffusion-based approaches, such as LDM and GSDM, generally achieve smoother visual reconstruction; however, the restored glyph appearances may still differ from the original glyph structures in some cases. PixelHacker further improves document-level restoration quality \cite{Xu2025}; however, such reconstruction-based approaches may still face challenges when distinguishing visually similar historical glyphs. In contrast, CRGR performs reference-based glyph restoration by retrieving intact glyph instances from a gallery of historical documents and verifying candidate consistency with a frozen vision-language model. Instead of synthesizing missing pixels, CRGR directly restores corrupted regions using verified reference glyphs. This strategy effectively preserves original glyph structures, printed stroke characteristics, and document layout while avoiding incorrect character reconstruction.

\begin{table*}
\centering
\small
\setlength{\tabcolsep}{8.0pt}
\begin{tabular}{lccccc | ccccc}
\toprule
& \multicolumn{5}{c}{Masked Regions} & \multicolumn{5}{c}{Intact Regions} \\
Method
& GRA$\uparrow$ & CCR$\downarrow$ & GSS$\uparrow$ & GSS$_m$$\uparrow$ & GSS$_h$$\uparrow$
& GRA$\uparrow$ & CCR$\downarrow$ & GSS$\uparrow$ & GSS$_m$$\uparrow$ & GSS$_h$$\uparrow$ \\
\midrule
LDM
& 0.9354 & 0.0456 & 0.9966 & 0.9969 & 0.9959
& 0.9439 & 0.0448 & 0.9974 & 0.9976 & 0.9970 \\

LaMa
& 0.8992 & 0.0684 & 0.9974 & 0.9977 & 0.9967
& 0.9589 & 0.0383 & \textbf{0.9997} & \textbf{0.9998} & \textbf{0.9997} \\

AOT-GAN
& 0.8764 & 0.1008 & 0.9963 & 0.9965 & 0.9957
& 0.9418 & 0.0428 & 0.9970 & 0.9973 & 0.9967 \\

GSDM
& 0.5627 & 0.0932 & 0.9971 & 0.9973 & 0.9966
& 0.5910 & 0.0688 & 0.9981 & 0.9982 & 0.9980 \\

PixelHacker
& \textbf{0.9487} & \textbf{0.0380} & 0.9967 & 0.9970 & 0.9959
& 0.9283 & 0.0452 & 0.9966 & 0.9971 & 0.9960 \\

CRGR (Ours)
& 0.9354 & 0.0456 & \textbf{0.9986} & \textbf{0.9989} & \textbf{0.9979}
& \textbf{0.9600} & \textbf{0.0377} & \textbf{0.9997} & \textbf{0.9998} & \textit{0.9996} \\
\bottomrule
\end{tabular}
\caption{Quantitative comparison of glyph-oriented restoration performance on masked and intact regions. }
\label{tab:glyph_metrics}
\end{table*}

\subsubsection{Quantitative Comparison.} 

Since historical document restoration requires preserving glyph structures and textual contents rather than merely reconstructing pixel appearances, we primarily evaluate all methods using the proposed glyph-oriented metrics. The results on masked and intact regions are summarized in Table~\ref{tab:glyph_metrics}. Here, GSS$_m$ and GSS$_h$ denote structural similarity evaluated on Manchu and Han glyph subsets, respectively.

On masked regions, CRGR achieves the highest GSS score of 0.9986, together with the best GSS$_m$ and GSS$_h$ scores among all compared methods. This result suggests that reference-based restoration can better preserve authentic glyph structures by directly selecting existing glyph instances rather than synthesizing missing contents. Although PixelHacker obtains slightly better GRA and CCR on masked regions, its lower structural similarity suggests that category-level correctness does not necessarily guarantee complete glyph structure preservation.

\begin{table}[b]
\centering
\small
\setlength{\tabcolsep}{1.3mm}
\begin{tabular}{lccccc}
\toprule
Method & PSNR$\uparrow$ & SSIM$\uparrow$ & LPIPS$\downarrow$ & RMSE$\downarrow$ & F1$\uparrow$ \\
\midrule
LDM & \textbf{34.671} & 0.996 & 0.004 & \textbf{4.875} & \textbf{0.999} \\
LaMa & 33.396 & \textbf{0.997} & 0.004 & 5.707 & \textbf{0.999} \\
AOT-GAN & 22.514 & 0.874 & 0.327 & 19.144 & 0.987 \\
GSDM & 20.464 & 0.936 & 0.123 & 25.787 & 0.987 \\
PixelHacker & 18.954 & 0.788 & 0.383 & 28.839 & 0.975 \\
CRGR (Ours) & 31.982 & 0.994 & \textbf{0.002} & 7.135 & \textbf{0.999} \\
\bottomrule
\end{tabular}
\caption{Quantitative comparison of restoration performance. }
\label{tab:visual_metrics}
\end{table}

On intact regions, CRGR achieves the best overall glyph-oriented performance, obtaining the highest GRA (0.9600), the lowest CCR (0.0377), and competitive GSS performance. In contrast, generation-based methods may introduce modifications beyond corrupted areas due to their image reconstruction nature. By restricting restoration to verified damaged glyph regions, CRGR better preserves the original document contents while avoiding over-restoration.

For completeness, conventional pixel-level metrics are also reported in Table~\ref{tab:visual_metrics}. Since these metrics mainly measure low-level appearance similarity, they cannot fully reflect glyph identity preservation and structural consistency. Nevertheless, CRGR achieves competitive visual quality with the lowest LPIPS score, while its main advantage is demonstrated by the glyph-oriented evaluation. This observation further suggests that pixel-level reconstruction quality may not fully reflect glyph correctness in historical document restoration, where semantic fidelity is often more important than appearance similarity alone.

\begin{table}[t]
\centering
\setlength{\tabcolsep}{0.5mm}
\small
\begin{tabular}{lccccc}
\toprule
Variant & GRA$\uparrow$ & CCR$\downarrow$ & GSS$\uparrow$ & GSS$_m$$\uparrow$ & GSS$_h$$\uparrow$ \\
\midrule
ConvNeXt & 0.6186 & 0.2542 & 0.9977 & 0.9979 & 0.9969 \\
Swin V2 & \underline{0.6356} & \underline{0.2288} & \underline{0.9979} & \underline{0.9980} & 0.9973 \\
ConvNeXt + Qwen-VL & 0.5763 & 0.2458 & 0.9976 & 0.9975 & \textbf{0.9980} \\
ConvNeXt + InternVL & 0.5678 & 0.2542 & 0.9976 & 0.9975 & \textbf{0.9980} \\
Swin V2 + Qwen-VL & \underline{0.6356} & 0.2373 & \textbf{0.9980} & \textbf{0.9981} & \underline{0.9976} \\
Swin V2 + InternVL & \textbf{0.6610} & \textbf{0.2203} & \underline{0.9979} & \underline{0.9980} & 0.9972 \\
\bottomrule
\end{tabular}
\caption{Ablation study of CRGR variants. \textbf{Bold}: best result; \underline{underlined}: second-best result.}
\label{tab:ablation}
\end{table}

\begin{figure*}
\centering
\includegraphics[width=\linewidth]{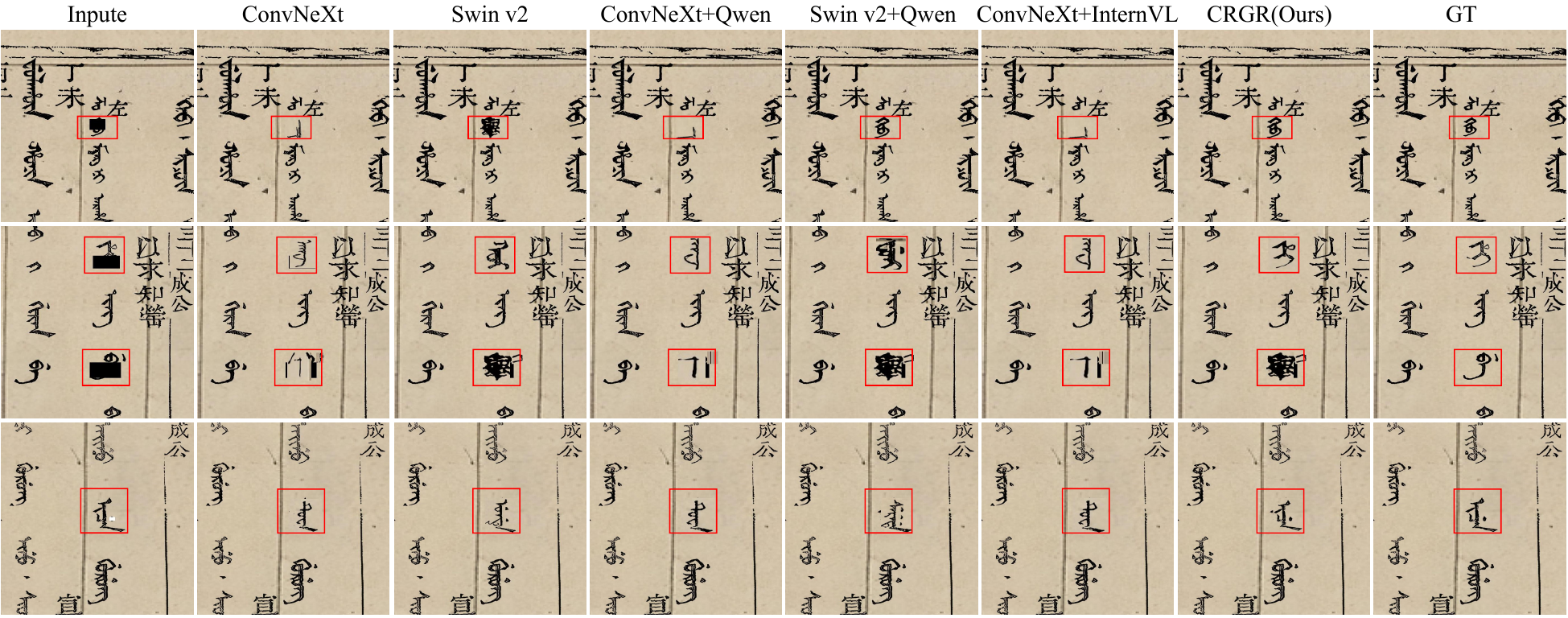}
\caption{Qualitative comparison of different CRGR variants.}
\label{fig:ablation}
\end{figure*}

\subsection{Ablation Study}
We perform ablation studies to analyze the contribution of each component in CRGR, including the glyph retrieval backbone, VLM-based verification, and the candidate number used for verification. All experiments are conducted under the same evaluation protocol, and results are reported using the proposed glyph-oriented metrics.

\begin{table}[b]
\centering
\small
\begin{tabular}{c ccccc}
\toprule
$K$ & GRA$\uparrow$ & CCR$\downarrow$ & GSS$\uparrow$ & GSS$_m$$\uparrow$ & GSS$_h$$\uparrow$\\
\midrule
$K=1$ & 0.6356 & 0.2288 & \underline{0.9979} & \underline{0.9980} & \underline{0.9973}\\
$K=3$ & 0.6356 & 0.2288 & \textbf{0.9980} & \textbf{0.9981} & \underline{0.9975}\\
$K=5$ & \textbf{0.6610} & \underline{0.2203} & \underline{0.9979} & \underline{0.9980} & 0.9972\\
$K=8$ & 0.6271 & 0.2373 & \underline{0.9979} & \textbf{0.9981} & 0.9972\\
$K=10$ & \underline{0.6525} & \textbf{0.2119} & \underline{0.9979} & \textbf{0.9981} & 0.9972\\
\bottomrule
\end{tabular}
\caption{Effect of candidate number $K$ in VLM verification. \textbf{Bold}: best result; \underline{underlined}: second-best result. }
\label{tab:topk}
\end{table}

\subsubsection{Effect of Glyph Retrieval Backbone.}
We analyze the effect of visual encoders for coarse glyph retrieval under the retrieval-only setting, we compare ConvNeXt \citep{Liu2022} and Swin V2 as representative modern visual backbones. As shown in Table~\ref{tab:ablation}, Swin V2 outperforms ConvNeXt by increasing GRA from 0.6186 to 0.6356 and reducing CCR from 0.2542 to 0.2288. Despite comparable GSS results, Swin V2 achieves better glyph structural consistency, demonstrating that hierarchical features are more discriminative for visually similar historical glyphs.

\begin{figure}[t]
\centering
\includegraphics[width=\linewidth]{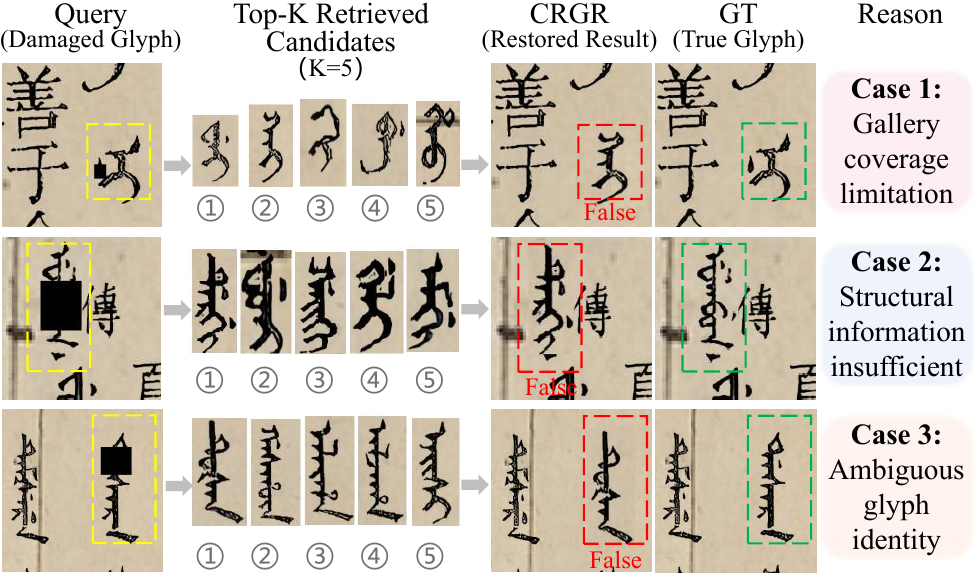}
\caption{Failure cases of CRGR caused by limited reference coverage, severe degradation, and glyph ambiguity.}
\label{fig:failure_cases}
\end{figure}

\subsubsection{Effect of VLM-based Verification.}
We further evaluate the contribution of vision-language verification by comparing different VLM backbones, including Qwen-VL \citep{Bai2023} and InternVL. When integrated with Swin V2 retrieval, VLM verification further improves restoration performance when reliable retrieval candidates are provided, yielding the best GRA (0.6610) and CCR (0.2203). In comparison, VLM verification does not consistently improve performance when the retrieval quality is insufficient, as observed in the ConvNeXt-based variants. This indicates that VLM verification benefits from high-quality retrieval candidates, while inaccurate candidate retrieval may limit fine-grained structural reasoning.

This result suggests that the two stages are complementary: retrieval provides structurally relevant candidates, while VLM-based verification further improves candidate discrimination among visually similar glyphs.

\subsubsection{Effect of Candidate Number.}
We additionally analyze the impact of retrieved candidate number $K$ in the verification stage. As shown in Table~\ref{tab:topk}, increasing $K$ expands the candidate pool for VLM verification, yet excessive candidates may bring extra ambiguities. The results reveal a trade-off between candidate diversity and verification reliability. A small $K$ restricts the candidate space and risks excluding the ground-truth glyph, while an overly large $K$ introduces additional candidates and increases verification complexity for VLM discrimination.
CRGR achieves the highest GRA of 0.6610 with competitive structural consistency at $K=5$, providing a favorable balance between retrieval coverage and verification accuracy. Larger $K$ values improve CCR in some cases but do not consistently improve overall performance while increasing verification cost. Accordingly, we adopt $K=5$ as the default configuration to balance retrieval coverage, verification reliability, computational efficiency and restoration quality. These results further validate the efficacy of our coarse-to-fine retrieval strategy.

\subsection{Failure Analysis}
Although CRGR achieves strong restoration performance, failure cases may still occur in challenging scenarios where the reference evidence is insufficient for accurate glyph identification. We analyze representative failure cases to investigate the limitations of the proposed retrieval-guided paradigm. As shown in Fig.~\ref{fig:failure_cases}, failures can be categorized into three scenarios: (1) gallery coverage limitation, where the correct glyph is absent from the reference gallery and therefore cannot be retrieved; (2) insufficient structural information, where severe degradation removes discriminative glyph cues and makes candidate retrieval unreliable; and (3)ambiguous glyph identity, where visually similar glyphs introduce uncertainty during candidate verification. These cases indicate that CRGR is mainly limited by reference coverage and the availability of discriminative structural evidence, rather than the restoration mechanism itself.

\section{Conclusion}
This paper presented CRGR, a retrieval-guided glyph-aware restoration framework for low-resource historical document restoration. Instead of treating restoration as unconstrained pixel generation, CRGR reformulates the problem as coarse-to-fine glyph retrieval followed by VLM-based candidate verification. This formulation enables identity-aware and structure-preserving restoration using authentic historical references rather than synthesized glyphs. Experimental results demonstrate that the proposed retrieval-and-verification paradigm improves glyph restoration fidelity while preserving intact document contents. More broadly, our study suggests that low-resource historical document restoration should move beyond unconstrained pixel generation toward retrieval-guided semantic restoration, where authentic references and multimodal reasoning jointly improve glyph structural preservation across diverse ancient writing low-resource historical document restoration scenarios.

\bibliography{aaai2027}

@article{Pal2017,
    author = {Pal, Kazim and Avery, Nicola and Boston, Pete and Campagnolo, Alberto and De Stefani, Caroline and Matheson-Pollock, Helen and Panozzo, Daniele and Payne, Matthew and Schüller, Christian and Sanderson, Chris and Scott, Chris and Smith, Philippa and Smither, Rachael and Sorkine-Hornung, Olga and Stewart, Ann and Stewart, Emma and Stewart, Patricia and Terras, Melissa and Walsh, Bernadette and Ward, Laurence and Yamada, Liz and Weyrich, Tim},
    title = {Digitally reconstructing the Great Parchment Book: 3D recovery of fire-damaged historical documents},
    journal = {Digital Scholarship in the Humanities},
    volume = {32},
    number = {4},
    pages = {887-917},
    year = {2017},
    month = {12},
    issn = {2055-7671},
    doi = {10.1093/llc/fqw057},
    url = {https://doi.org/10.1093/llc/fqw057},
    eprint = {https://academic.oup.com/dsh/article-pdf/32/4/887/21343185/fqw057.pdf},
}

@INPROCEEDINGS{Suvorov2022,
  author={Suvorov, Roman and Logacheva, Elizaveta and Mashikhin, Anton and Remizova, Anastasia and Ashukha, Arsenii and Silvestrov, Aleksei and Kong, Naejin and Goka, Harshith and Park, Kiwoong and Lempitsky, Victor},
  booktitle={2022 IEEE/CVF Winter Conference on Applications of Computer Vision (WACV)}, 
  title={Resolution-robust Large Mask Inpainting with Fourier Convolutions}, 
  year={2022},
  volume={},
  number={},
  pages={3172-3182},
  doi={10.1109/WACV51458.2022.00323}}

@ARTICLE{Zeng2021,
  author={Zeng, Yanhong and Fu, Jianlong and Chao, Hongyang and Guo, Baining},
  journal={IEEE Transactions on Visualization and Computer Graphics}, 
  title={Aggregated Contextual Transformations for High-Resolution Image Inpainting}, 
  year={2023},
  volume={29},
  number={7},
  pages={3266-3280},
  doi={10.1109/TVCG.2022.3156949}}

@InProceedings{Ruan2026,
    author    = {Ruan, Wentao and Li, Xinhui and Cheng, Zhan and Fan, Cunhang and Tian, Libao and Lv, Zhao},
    title     = {LWTformer: A Detail-Aware, Learnable Wavelet-Transformer for Ancient Chinese Character Image Restoration},
    booktitle = {Proceedings of the IEEE/CVF Conference on Computer Vision and Pattern Recognition (CVPR) Findings},
    month     = {June},
    year      = {2026},
    pages     = {4971-4981}
}

@article{Kim2025,
  title={Unified Diffusion Transformer for High-fidelity Text-Aware Image Restoration},
  author={Jin Hyeon Kim and Paul Hyunbin Cho and Claire Kim and Jaewon Min and Jaeeun Lee and Jihye Park and Yeji Choi and Seungryong Kim},
  journal={ArXiv},
  year={2025},
  volume={abs/2512.08922},
  url={https://api.semanticscholar.org/CorpusID:283711857}
}

@article{Zhu2024,
  title={Text Image Inpainting via Global Structure-Guided Diffusion Models},
  author={Shipeng Zhu and Pengfei Fang and Chenjie Zhu and Zuoyan Zhao and Qiang Xu and Hui Xue},
  journal={ArXiv},
  year={2024},
  volume={abs/2401.14832},
  url={https://api.semanticscholar.org/CorpusID:267301109}
}

@inproceedings{Li2024,
author = {Li, Haolong and Du, Chenghao and Jiang, Ziheng and Zhang, Yifan and Ma, Jiawei and Ye, Chen},
title = {Towards automated chinese ancient character restoration: a diffusion-based method with a new dataset},
year = {2024},
isbn = {978-1-57735-887-9},
publisher = {AAAI Press},
url = {https://doi.org/10.1609/aaai.v38i4.28090},
doi = {10.1609/aaai.v38i4.28090},
articleno = {342},
numpages = {9},
series = {AAAI'24/IAAI'24/EAAI'24}
}

@misc {Zheng2022,
	Title = {EA-GAN: Ancient books text restoration model based on example attention},
	Author = {Zheng, Wenjun and Su, Benpeng and Feng, Ruiqi and Peng, Xihua and Chen, Shanxiong},
	DOI = {10.21203/rs.3.rs-2131172/v1},
	Publisher = {Research Square},
	Year = {2022},
	URL = {https://doi.org/10.21203/rs.3.rs-2131172/v1},
}

@inproceedings{Duan2024,
    title = "Restoring Ancient Ideograph: A Multimodal Multitask Neural Network Approach",
    author = "Duan, Siyu  and
      Wang, Jun  and
      Su, Qi",
    editor = "Calzolari, Nicoletta  and
      Kan, Min-Yen  and
      Hoste, Veronique  and
      Lenci, Alessandro  and
      Sakti, Sakriani  and
      Xue, Nianwen",
    booktitle = "Proceedings of the 2024 Joint International Conference on Computational Linguistics, Language Resources and Evaluation (LREC-COLING 2024)",
    month = may,
    year = "2024",
    address = "Torino, Italia",
    publisher = "ELRA and ICCL",
    url = "https://aclanthology.org/2024.lrec-main.1221/",
    pages = "14005--14015",
}

@article{Liu2025,
  title={A structural information-guided cross-modal method for damaged inscription inpainting via vision-language models},
  author={Yunjing Liu and Erhu Zhang and Guangfeng Lin and Jinghong Duan},
  journal={npj Heritage Science},
  year={2025},
  volume={13},
  url={https://api.semanticscholar.org/CorpusID:281799939}
}

@article{Zhu2026,
  title={EpiAgent: An Agent-Centric System for Ancient Inscription Restoration},
  author={Shipeng Zhu and Ang Chen and Na Nie and Pengfei Fang and Min-Ling Zhang and Hui Xue},
  journal={ArXiv},
  year={2026},
  volume={abs/2604.09367},
  url={https://api.semanticscholar.org/CorpusID:287352073}
}

@Article{Shen2026,
AUTHOR = {Shen, Wenhan and Xu, Yubo and Zhang, Chaoqing and Yan, Juan and Wang, Shibin},
TITLE = {A Three-Stage Generative Adversarial Image Inpainting Framework for Broken-Stroke Restoration in Historical Rubbings: A Case Study on Oracle Bone Rubbings},
JOURNAL = {Applied Sciences},
VOLUME = {16},
YEAR = {2026},
NUMBER = {9},
ARTICLE-NUMBER = {4306},
URL = {https://www.mdpi.com/2076-3417/16/9/4306},
ISSN = {2076-3417},
DOI = {10.3390/app16094306}
}

@article{Lu2025,
  title={TextDoctor: Unified Document Image Inpainting via Patch Pyramid Diffusion Models},
  author={Wanglong Lu and Li Su and Jingjing Zheng and Vin'icius Veloso de Melo and Farzaneh Shoeleh and John Hawkin and Terrence Tricco and Hanli Zhao and Xianta Jiang},
  journal={ArXiv},
  year={2025},
  volume={abs/2503.04021},
  url={https://api.semanticscholar.org/CorpusID:276813584}
}

@article{Chen2022,
author = {Chen, Shanxiong and Yang, Ye and Liu, Xuxin and Zhu, Shiyu},
title = {Dual Discriminator GAN: Restoring Ancient Yi Characters},
year = {2022},
issue_date = {July 2022},
publisher = {Association for Computing Machinery},
address = {New York, NY, USA},
volume = {21},
number = {4},
issn = {2375-4699},
url = {https://doi.org/10.1145/3490031},
doi = {10.1145/3490031},
journal = {ACM Trans. Asian Low-Resour. Lang. Inf. Process.},
month = mar,
articleno = {66},
numpages = {23}
}

@article{Purkayastha2026,
  title={DocRevive: A Unified Pipeline for Document Text Restoration},
  author={Kunal Purkayastha and Ayan Banerjee and Josep Llad{\'o}s and Umapada Pal},
  journal={ArXiv},
  year={2026},
  volume={abs/2604.10077},
  url={https://api.semanticscholar.org/CorpusID:287425485}
}

@inproceedings{Zhang2026,
    title = "Draft, Verify, Restore: Self-Refining Historical Inscription Restoration with a Unified {MLLM}",
    author = "Zhang, Yuyi  and
      Liu, Junle  and
      Zhang, Peirong  and
      Liu, Jianliang  and
      Yang, Zhenhua  and
      Jin, Lianwen",
    editor = "Liakata, Maria  and
      Moreira, Viviane P.  and
      Zhang, Jiajun  and
      Jurgens, David",
    booktitle = "Proceedings of the 64th Annual Meeting of the {A}ssociation for {C}omputational {L}inguistics (Volume 1: Long Papers)",
    month = jul,
    year = "2026",
    address = "San Diego, California, United States",
    publisher = "Association for Computational Linguistics",
    url = "https://aclanthology.org/2026.acl-long.1254/",
    doi = "10.18653/v1/2026.acl-long.1254",
    pages = "27216--27231",
    ISBN = "979-8-89176-390-6"
}

@INPROCEEDINGS{Pathak2016,
  author={Pathak, Deepak and Krähenbühl, Philipp and Donahue, Jeff and Darrell, Trevor and Efros, Alexei A.},
  booktitle={2016 IEEE Conference on Computer Vision and Pattern Recognition (CVPR)}, 
  title={Context Encoders: Feature Learning by Inpainting}, 
  year={2016},
  volume={},
  number={},
  pages={2536-2544},
  doi={10.1109/CVPR.2016.278}}

@INPROCEEDINGS{Yu2019,
  author={Yu, Jiahui and Lin, Zhe and Yang, Jimei and Shen, Xiaohui and Lu, Xin and Huang, Thomas},
  booktitle={2019 IEEE/CVF International Conference on Computer Vision (ICCV)}, 
  title={Free-Form Image Inpainting With Gated Convolution}, 
  year={2019},
  volume={},
  number={},
  pages={4470-4479},
  doi={10.1109/ICCV.2019.00457}}

@INPROCEEDINGS{Liu2021,
  author={Liu, Ze and Lin, Yutong and Cao, Yue and Hu, Han and Wei, Yixuan and Zhang, Zheng and Lin, Stephen and Guo, Baining},
  booktitle={2021 IEEE/CVF International Conference on Computer Vision (ICCV)}, 
  title={Swin Transformer: Hierarchical Vision Transformer using Shifted Windows}, 
  year={2021},
  volume={},
  number={},
  pages={9992-10002},
  doi={10.1109/ICCV48922.2021.00986}}

@article{Radford2021,
  author = {Alec Radford and Jong Wook Kim andChris Hallacy and Aditya Ramesh and Gabriel Goh and Sandhini Agarwal and Girish Sastry and Amanda Askell and Pamela Mishkin and Jack Clark and Gretchen Krueger and Ilya Sutskever},
  title = {Learning Transferable Visual Models From Natural Language Supervision},
  journal = {CoRR},
  volume  = {abs/2103.00020},
  year = {2021},
  url = {https://arxiv.org/abs/2103.00020},
  eprinttype = {arXiv},
  eprint = {2103.00020},
  bibsource = {dblp computer science bibliography, https://dblp.org}
}

@misc{Chen2024,
      title={InternVL: Scaling up Vision Foundation Models and Aligning for Generic Visual-Linguistic Tasks}, 
      author={Zhe Chen and Jiannan Wu and Wenhai Wang and Weijie Su and Guo Chen and Sen Xing and Muyan Zhong and Qinglong Zhang and Xizhou Zhu and Lewei Lu and Bin Li and Ping Luo and Tong Lu and Yu Qiao and Jifeng Dai},
      year={2024},
      eprint={2312.14238},
      archivePrefix={arXiv},
      primaryClass={cs.CV},
      url={https://arxiv.org/abs/2312.14238}, 
}

@INPROCEEDINGS{Liu2022,
  author={Liu, Zhuang and Mao, Hanzi and Wu, Chao-Yuan and Feichtenhofer, Christoph and Darrell, Trevor and Xie, Saining},
  booktitle={2022 IEEE/CVF Conference on Computer Vision and Pattern Recognition (CVPR)}, 
  title={A ConvNet for the 2020s}, 
  year={2022},
  volume={},
  number={},
  pages={11966-11976},
  doi={10.1109/CVPR52688.2022.01167}}

@misc{Bai2023,
      title={Qwen-VL: A Versatile Vision-Language Model for Understanding, Localization, Text Reading, and Beyond}, 
      author={Jinze Bai and Shuai Bai and Shusheng Yang and Shijie Wang and Sinan Tan and Peng Wang and Junyang Lin and Chang Zhou and Jingren Zhou},
      year={2023},
      eprint={2308.12966},
      archivePrefix={arXiv},
      primaryClass={cs.CV},
      url={https://arxiv.org/abs/2308.12966}, 
}

@misc{Rombach2021,
      title={High-Resolution Image Synthesis with Latent Diffusion Models}, 
      author={Robin Rombach and Andreas Blattmann and Dominik Lorenz and Patrick Esser and Björn Ommer},
      year={2021},
      eprint={2112.10752},
      archivePrefix={arXiv},
      primaryClass={cs.CV}
}

@article{Xu2025,
  title={PixelHacker: Image Inpainting with Structural and Semantic Consistency},
  author={Ziyang Xu and Kangsheng Duan and Xi Shen and Zhifeng Ding and Wenyu Liu and Xiaohui Ruan and Xiaoxin Chen and Xinggang Wang},
  journal={ArXiv},
  year={2025},
  volume={abs/2504.20438},
  url={https://api.semanticscholar.org/CorpusID:278200760}
}


\end{document}